\documentclass[oribibl]{llncs}

\usepackage{cite}
  \usepackage[pdftex]{graphicx}
  \usepackage{epstopdf}
    \DeclareGraphicsExtensions{.eps,.pdf,.jpeg,.png}
\usepackage{array}
\usepackage{mdwtab}
\usepackage{multirow}
\usepackage{amsmath}
\usepackage[caption=false,font=footnotesize,farskip=4px,captionskip=-8px]{subfig}

\begin{document}

\title{Fitness Landscape-Based Characterisation\\of Nature-Inspired Algorithms \\
{\small{To appear in the Proceedings of ICANNGA'13, 11th International Conference on Adaptive and Natural Computing Algorithms, Springer.}}}
\author{Matthew~Crossley \and Andy~Nisbet \and Martyn~Amos
\thanks{Matthew Crossley is supported by a Ph.D. studentship from the Dalton Research Institute, MMU. The authors thank David Corne for useful discussions.}}
\institute{School of Computing, Mathematics and Digital Technology,\\Manchester Metropolitan University, Manchester M15GD, UK
\email{\{m.crossley,a.nisbet,m.amos\}@mmu.ac.uk}}
%\thanks{The authors are with the School of Computing, Mathematics and Digital Technology, Manchester Metropolitan University,
%Manchester M1 5GD, United Kingdom (e-mail: m.crossley@mmu.ac.uk; a.nisbet@mmu.ac.uk; m.amos@mmu.ac.uk).}}
\maketitle

\begin{abstract}
A significant challenge in nature-inspired algorithmics is the identification of specific characteristics of problems that make them harder (or easier) to solve using specific methods. The hope is that, by identifying these characteristics, we may more easily predict which algorithms are best-suited to problems sharing certain features. Here, we approach this problem using fitness landscape analysis. Techniques already exist for  measuring the ``difficulty" of specific landscapes, but these are often designed solely with evolutionary algorithms in mind, and are generally specific to discrete optimisation.  In this paper we develop an approach for comparing a wide range of continuous optimisation algorithms. Using a fitness landscape generation technique, we compare six different nature-inspired algorithms and identify which methods perform best on landscapes exhibiting specific features.
\end{abstract}

%\IEEEpeerreviewmaketitle

\section{Introduction}

Inspired by the foundational work of Wolpert and Macready \cite{Wolpert1997}, practitioners have long sought to better understand the relationship between  problems and solution methods (i.e., algorithms). Here, we are particularly interested in the question of which algorithm is {\it best-suited} to a particular problem, and the process of addressing this has been described by some as a ``black-art" \cite{Woodward2010}.

Although theoretical studies in this area have yielded useful results, the {\it experimental analysis} of algorithms is receiving increasing attention. As Morgan and Gallagher point out \cite{Morgan2011}, this approach is {\it scalable} in that it readily admits newly-described algorithms, and it is now an area of research that is supported by a number of high-profile competitions and libraries of benchmark test problems.

The fundamental properties of a problem's {\it search landscape} underpin much work in experimental analysis, and the use of landscape/test case generators \cite{Gallagher2006,Jani2008,Morgan2011,Jin2004,Michalewicz2000} has been proposed as one way in which we might effectively assess algorithms against problem instances. 

In this paper we examine six different nature-inspired algorithms by testing them against a number of different randomized landscapes with several different properties (e.g., ruggedness). This gives a much richer picture of their relative strengths and weaknesses, compared to simply using the ``difficulty" of a landscape \cite{Jones1995}.

The rest of the paper is organized as follows: in Section ~\ref{previous} we give a brief overview of previous work, before describing our testing methodology in Section ~\ref{methodology}. We then present our experimental results in Section ~\ref{results}, before concluding in Section ~\ref{conclusion} with a discussion of our findings.

\section{Previous work}
\label{previous}

The use of algorithms inspired by physical or natural processes is now well-established in the field of optimisation \cite{Chiong2009}.
As the number of such algorithms grows year-on-year, there is a pressing need to better understand their properties, in order that practitioners may make informed decisions about which method is best-suited to a particular problem, under certain conditions. Although analytical methods have been successfully applied to nature-inspired methods \cite{Zhang2004} \cite{He2002}, their ``real world" applicability is not clear, as they often rely on significant assumptions and/or simplifications.

In what follows, we take an {\it experimental} approach\cite{Barr1995}  to studying the selected algorithms, using an established landscape generation technique \cite{Gallagher2006}. As Morgan and Gallagher observe, ``In a general sense, an algorithm can be expected to perform well if the assumptions that it makes, either explicit or implicit, are well-matched to the properties of the search landscape or solution space of a given problem or set of problems" \cite{Morgan2011}. We therefore seek to investigate the performance of several algorithms on a number of types of {\it fitness landscape} with specific properties or characteristics.  This approach is preferred by Hooker to the use of benchmark problems, because the latter ``differ in so many respects that it is rarely evident why some are harder than others, and they may yet fail to vary over parameters that are key determinants of performance. It is better generate problems in a controlled fashion... The goal is not to generate realistic problems, which random generation cannot do, but to generate several problem sets, each of which is homogeneous with respect to characteristics that are likely to affect performance" \cite{Hooker1995}.

The fitness landscape approach has been successfully applied to the study of various nature-inspired algorithms \cite{Freisleben2000,Tavares2008,Uludag2009}.  Indeed, to our knowledge, landscape analysis of nature-inspired algorithms has been largely {\it restricted} to evolutionary methods. In this paper we broaden this work {\it considerably}, by considering several classes of natural algorithms (social, evolutionary and physical). Overall, we study six different nature-inspired methods, as well as stochastic hill-climbing as a baseline algorithm. Our empirical approach is informed by previous work \cite{McGeoch1996} \cite{Eiben}, which emphasises the need to establish a rigorous framework for experimental algorithmics. In the next Section, we describe in detail our methodology.

\section{Methodology}
\label{methodology}

\subsection{Algorithm selection}

We select, for comparison, a number of nature-inspired algorithms that are commonly applied to continuous function optimisation. These may be classified \cite{Brabazon2006} as either {\it social}, {\it evolutionary} or {\it physical}. The social algorithms we select are Bacterial Foraging Optimisation Algorithm (BFOA) \cite{Passino2002}, Bees Algorithm (BA) \cite{Pham2006}, and Particle Swarm Optimisation (PSO) \cite{Kennedy1995}. The evolutionary algorithms selected are Genetic Algorithms (GA) \cite{Goldberg1989} and Evolution Strategies (ES) \cite{Back1993}, and physical algorithms are represented by Harmony Search (HS) \cite{Geem2001}. We also include random search (RS) and stochastic hill climbing (SHC) as ``baseline" algorithms.

We note that the references supplied above for each algorithm may serve simply as an example of their {\it application}, rather than their precise {\it implementation}. In terms of implementation, we heed the observation that ``Ideally, competing algorithms would be coded by the same expert programmer and run on the same test problems on the same computer configuration"  \cite{Barr1995}. With that in mind, we use only implementations provided by Brownlee to accompany \cite{Brownlee2011}.  The limited space available prevents a complete description of each algorithm, but full implementation details are in \cite{Brownlee2011}, which is freely available and contains the source code used here.

\subsection{Optimisation problem characteristics}

As Morgan and Gallagher explain \cite{Morgan2011}, their Max-Set of Gaussians (MSG) method \cite{Gallagher2006} is a ``randomised landscape generator that specifies test problems as a weighted sum of Gaussian functions. By specifying the number of Gaussians and the mean and covariance parameters for each component, a variety of test landscape instances can be generated. The topological properties of the landscapes are intuitively related to (and vary smoothly with) the parameters of the generator." By manipulating these parameters, we obtain landscapes with different {\it characteristics}. This allows us to investigate the performance of our selected algorithms on landscapes with different features, and to identify which characteristics pose the greatest challenge. As Morgan and Gallagher observe, ``Different problem types have their own characteristics, however it is usually the case that complementary insights into algorithm behaviour result from conducting larger experimental studies using a variety of different problem types" \cite{Morgan2011}. We now describe the different characteristics (corresponding to problem types) under study in this paper.

{\it Ruggedness} of a landscape is often linked to its difficulty \cite{Jones1995}, and factors affecting this include (1) the {\it number} of local optima  \cite{Horn1994}, and (2) {\it ratio} of the fitness value of local optima to the global optimal value \cite{Malan2009} \cite{Freisleben2000}.  Other significant factors concern (3) {\it dimensionality} \cite{Hendtlass2009} (that is, the number of variables in the objective function), (4) {\it boundary constraints} (that is, the limits imposed on the value of a variable) \cite{Kukkonen}, and (5) {\it smoothness} of each Gaussian curve (effectively the gradient) used to generate the landscape \cite{Beyer2002} - a smaller value indicates a smoother gradient. A summary of the ranges selected for each characteristic is given in Table \ref{featurespace}. 

		\renewcommand{\arraystretch}{1.3}
		\begin{table}[!t]
			\begin{center}
				\caption[]{A summary of the ranges selected for the characteristics in our fitness space ($F$)}
				\label{featurespace}
				\begin{tabular}[c]{p{4cm} c c c c }
					\hline\noalign{\smallskip}
					{\bf Characteristic} & {\bf Min} & {\bf Step} & {\bf Max} & {\bf Default} \\
					\noalign{\smallskip}
					\hline
					\noalign{\smallskip}
					Number of local optima & 0 & 1 & 9 & 3\\
					Ratio of local optima to global optimum & 0.1 & 0.2 & 0.9 & 0.5\\
					Dimensionality & 1 & 1 & 10 & 2\\
					Boundary constraints & 10 & 10 & 100 & 30\\
					Smoothness & 10 & 10 & 100 & 15\\
					\hline
				\end{tabular}
			\end{center}
		\end{table}

\subsection{Performance measurement}

In terms of {\it performance metrics}, we abstract away from algorithm-specific measures, due to the diverse range of methods selected. The following metrics are applied: {\bf (1) Accuracy}: We define this as the mean absolute error of the best solution found on a given set of landscape characteristics, over a number of runs ($\frac{1}{n}\displaystyle\sum\limits_{i=1}^n (x_{i} - \bar{x}$)) (where $X$ is the set of best solutions found, $n$ is the number of runs performed and $\bar{x}$ is the known optimum).  This is the most commonly-used assessment metric for optimisation algorithms \cite{Gallagher2006}.  The generation technique we use creates landscapes with a known global optimum, in this case zero. {\bf (2) Variance of final solutions}: A measure of variation in best solutions found across differently seeded runs. We use the standard deviation of the best solutions of all runs on a given set of landscape characteristics, defined as $ (\frac{1}{n-1}\displaystyle\sum\limits_{i=1}^n (x_{i} - \bar{x})^2)^{\frac{1}{2}}$ (where $X$ is our data set, $n$ is the size of the data set and $\bar{x}$ is the mean average). {\bf (3) Success rate}: We measure this as the frequency with which differently-seeded runs of an algorithm are able to find a solution within a specified distance from the optimum \cite{Elbeltagi2005}.  We keep the success tolerance relatively low (error less than 1.0$\times$10$^{-4}$) in order to ensure that we capture the change in success rate of algorithms which perform poorly.  
	
\subsection{Experimental setup}

In order to generate the landscapes, we used the Matlab code supplied with \cite{Gallagher2006}. 
All landscapes were generated using default parameters of three curves, two dimensions, 0.5 average ratio of local minima to global minimum, 30 units in each dimension with a smoothness coefficient of 15), with only the parameter under investigation changing for each experiment.
We ran each algorithm 100 times on each landscape in the set of landscapes generated for each particular characteristic value (when investigating smoothness, for example, we generated 10 different landscapes (smoothness = 10 \dots 100), and ran each algorithm 100 times on each landscape). 

{\it Parameterisation} of algorithms provides a significant challenge when evaluating performance.  Our aim is not to perform ``competitive testing" \cite{Hooker1995}, but to establish general performance {\it profiles} for different algorithms over different types of problem. As such, we use the so-called ``vanilla" implementation of each algorithm, with general-purpose settings taken from \cite{Gallagher2006}. Where an algorithm has a ``population size" parameter, we use a value of 50; where an algorithm has a ``range" or ``velocity" parameter, we use a value of 10.

{\it Termination criteria} were also standardised.  The most objective criterion is the number of objective function evaluations.  This means each algorithm has access to the same amount of information from the landscape, and the same amount of {\it feedback} on potential solutions.  Experimentally we determined that the selected algorithms generally converged within 20,000 objective function calculations, so this was used as the termination criterion. The code used for all algorithms, as well as datasets and the landscape generator, is available on request from the authors.

\section{Results}
\label{results}

\begin{figure}[]
			\centering \subfloat[Bees algorithm]{\includegraphics[height=0.20\textheight]{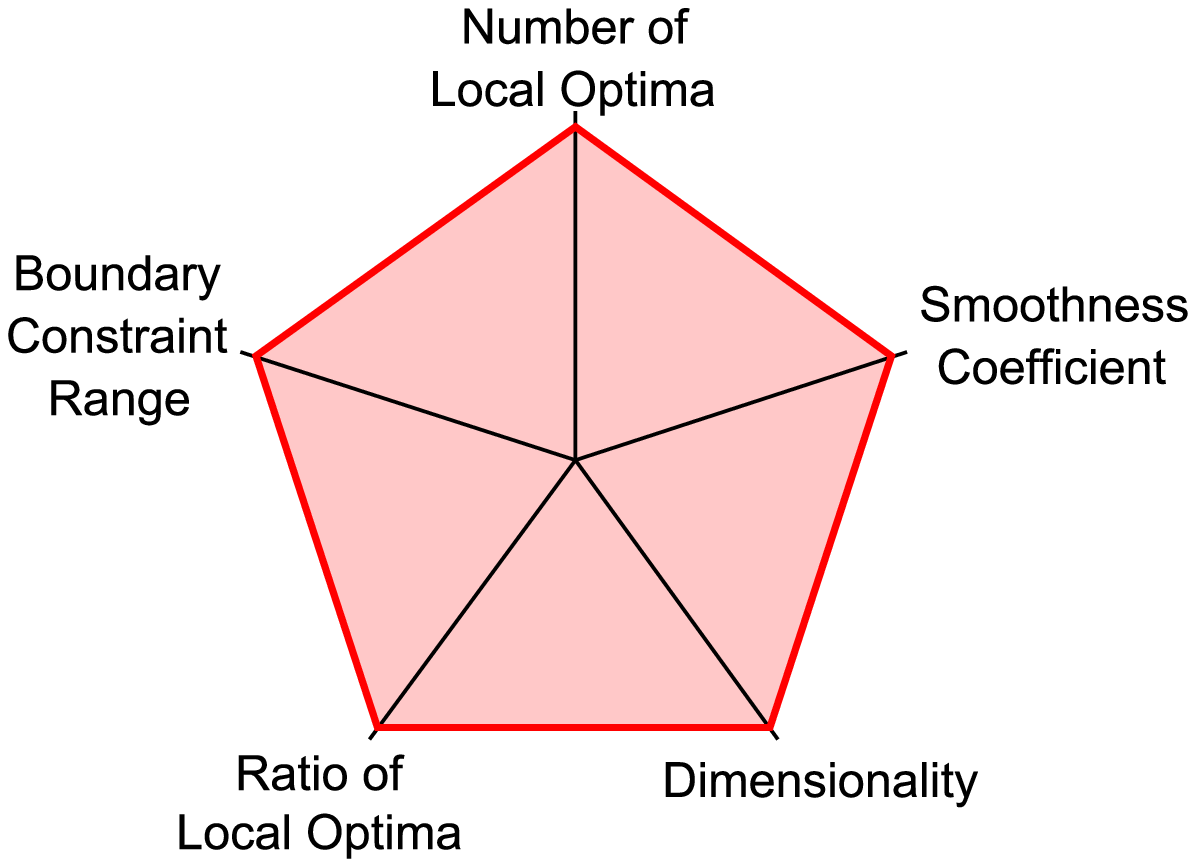}}%
			\hfil
			\subfloat[Bacterial foraging optimisation algorithm]{\includegraphics[height=0.20\textheight]{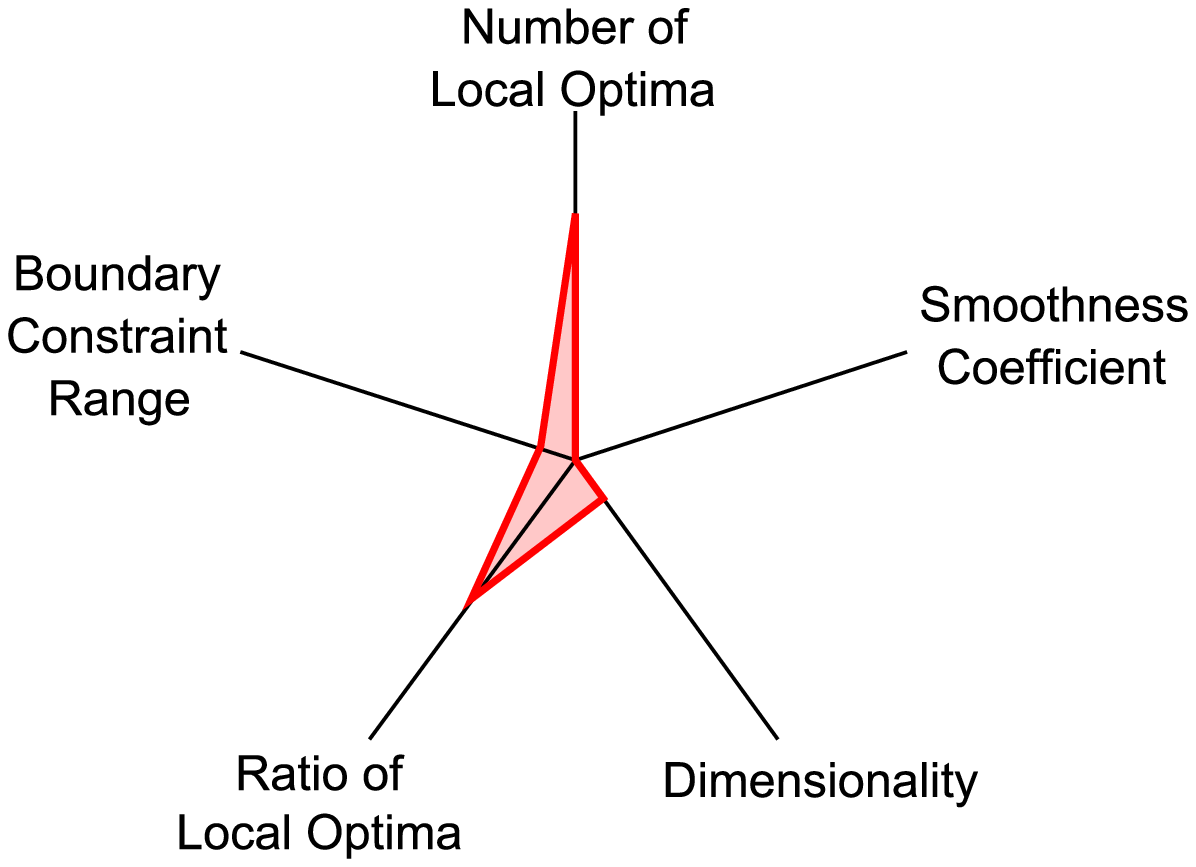}}%
			\hfil
			\subfloat[Evolution strategies]{\includegraphics[height=0.20\textheight]{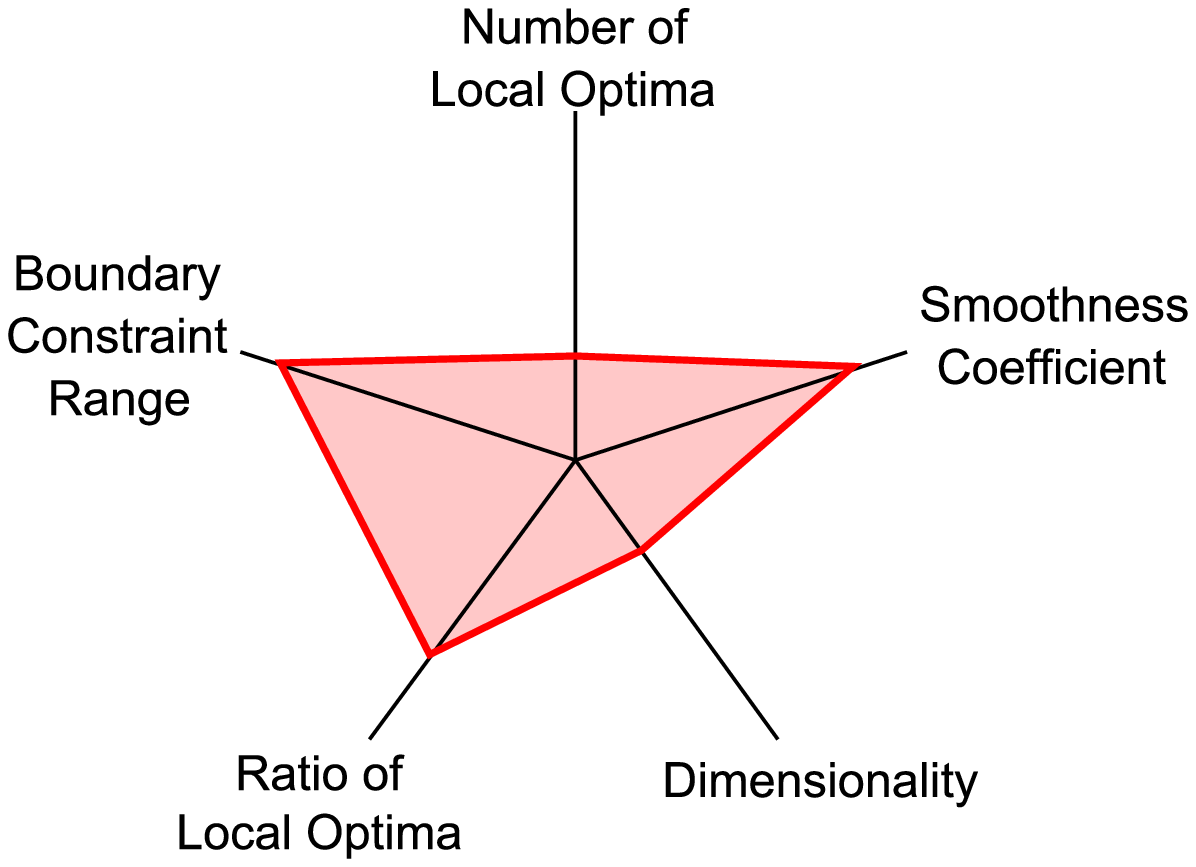}}%
			\hfil
			\subfloat[Genetic algorithm]{\includegraphics[height=0.2\textheight]{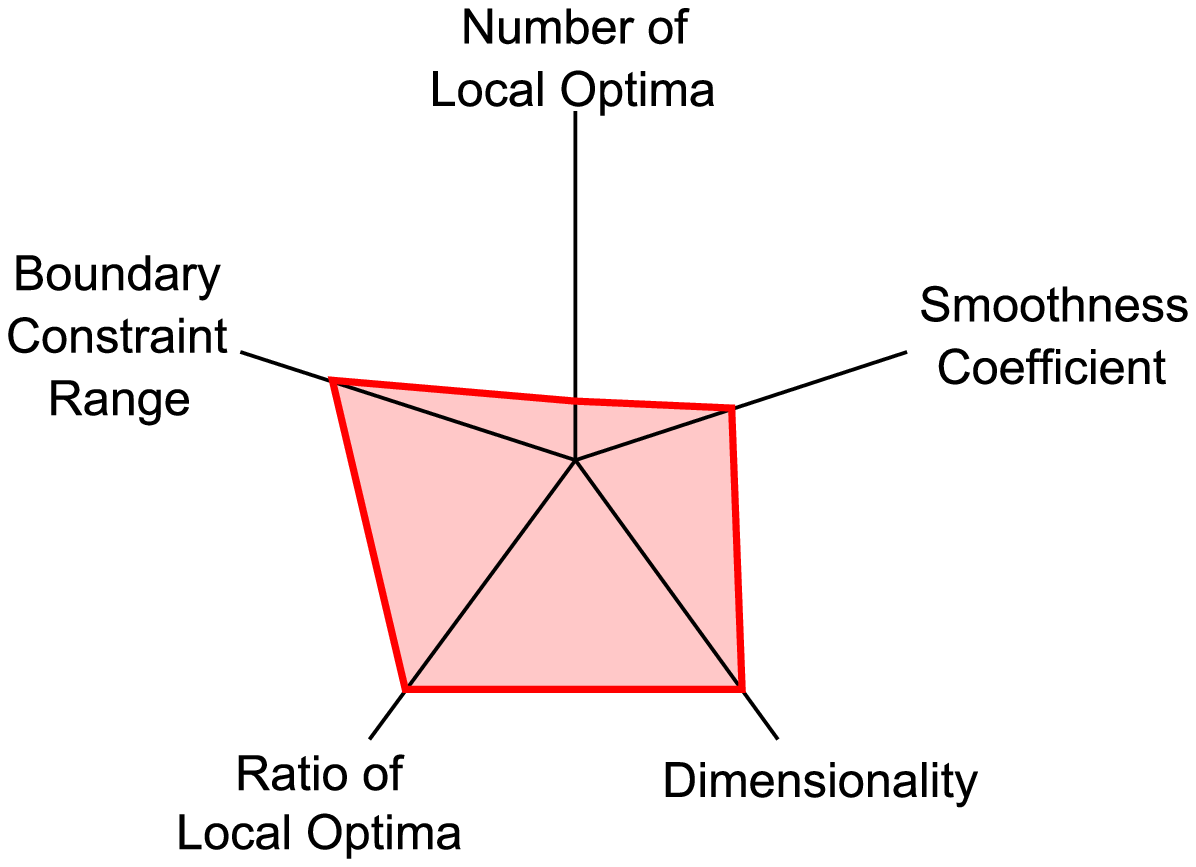}}%
			\hfil
			\subfloat[Harmony search]{\includegraphics[height=0.2\textheight]{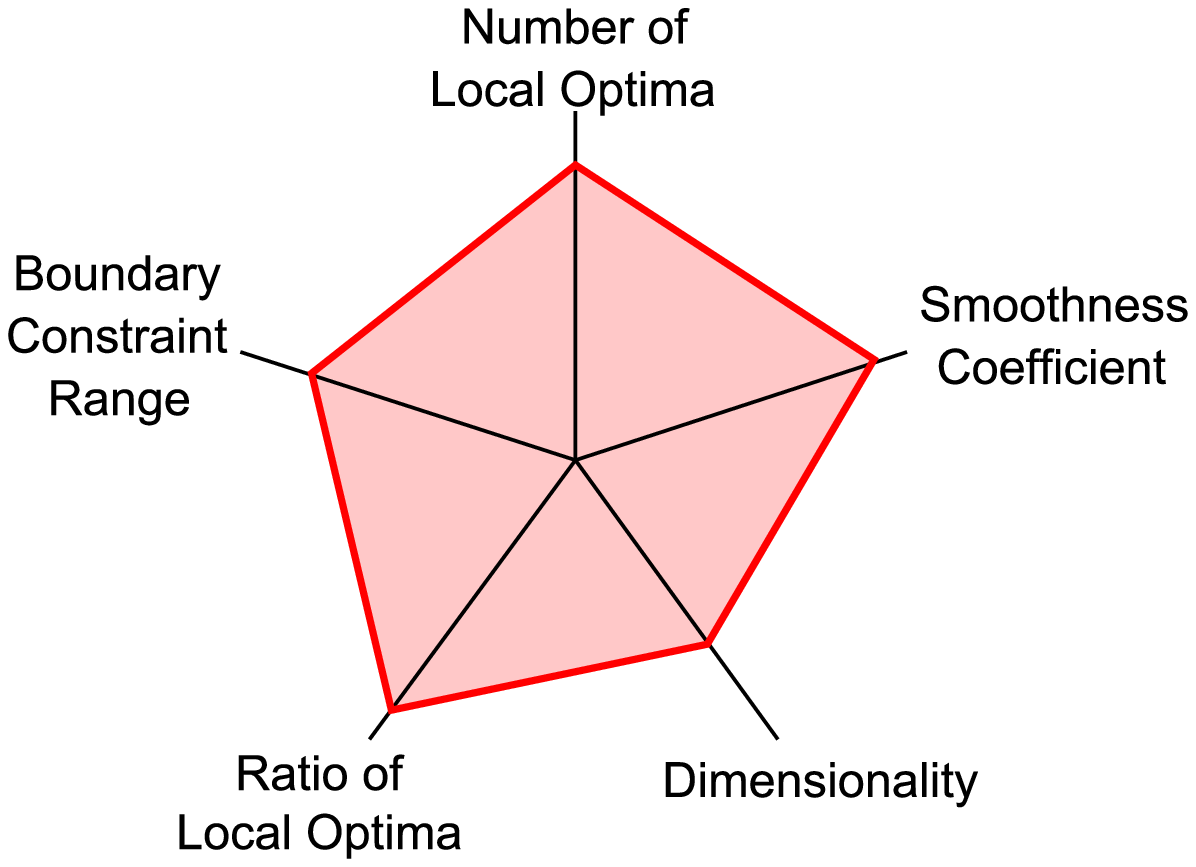}}%
			\hfil
			\subfloat[Particle swarm optimisation]{\includegraphics[height=0.2\textheight]{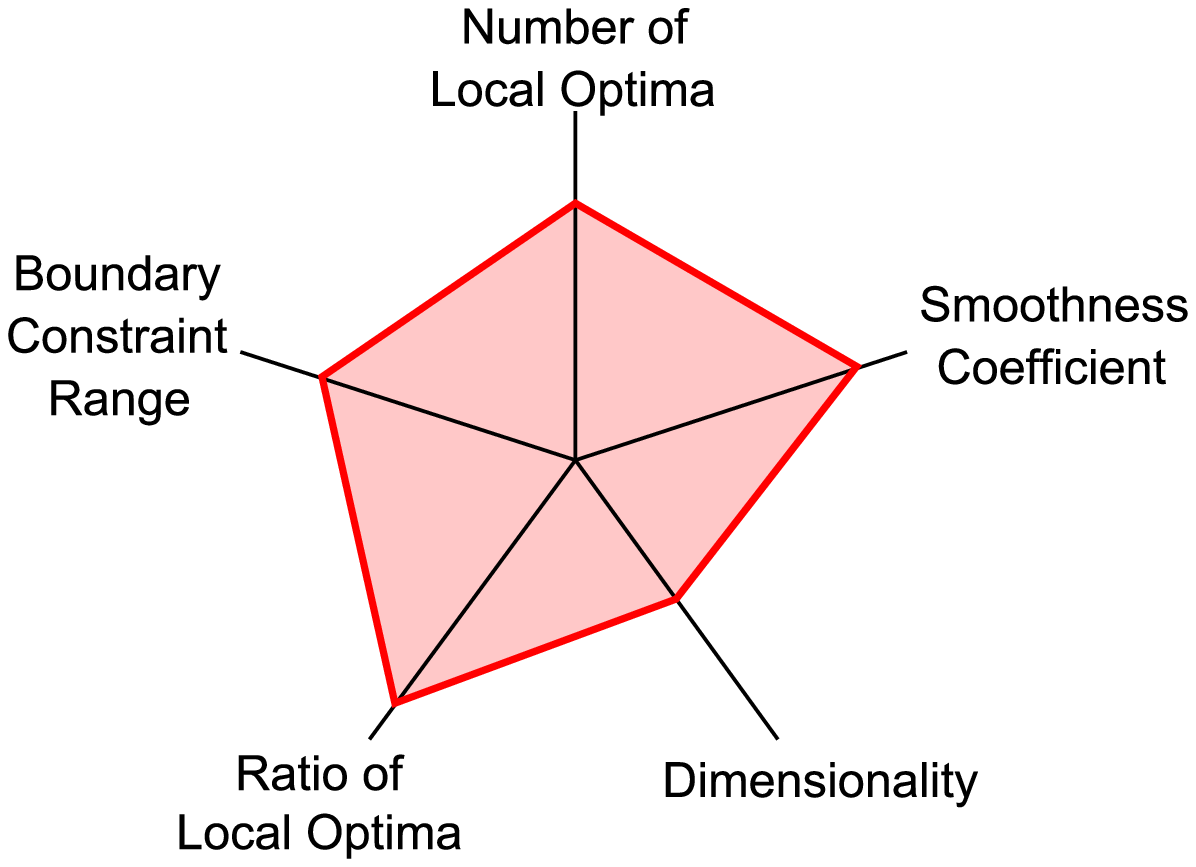}}%
			\hfil
			\subfloat[Random search]{\includegraphics[height=0.2\textheight]{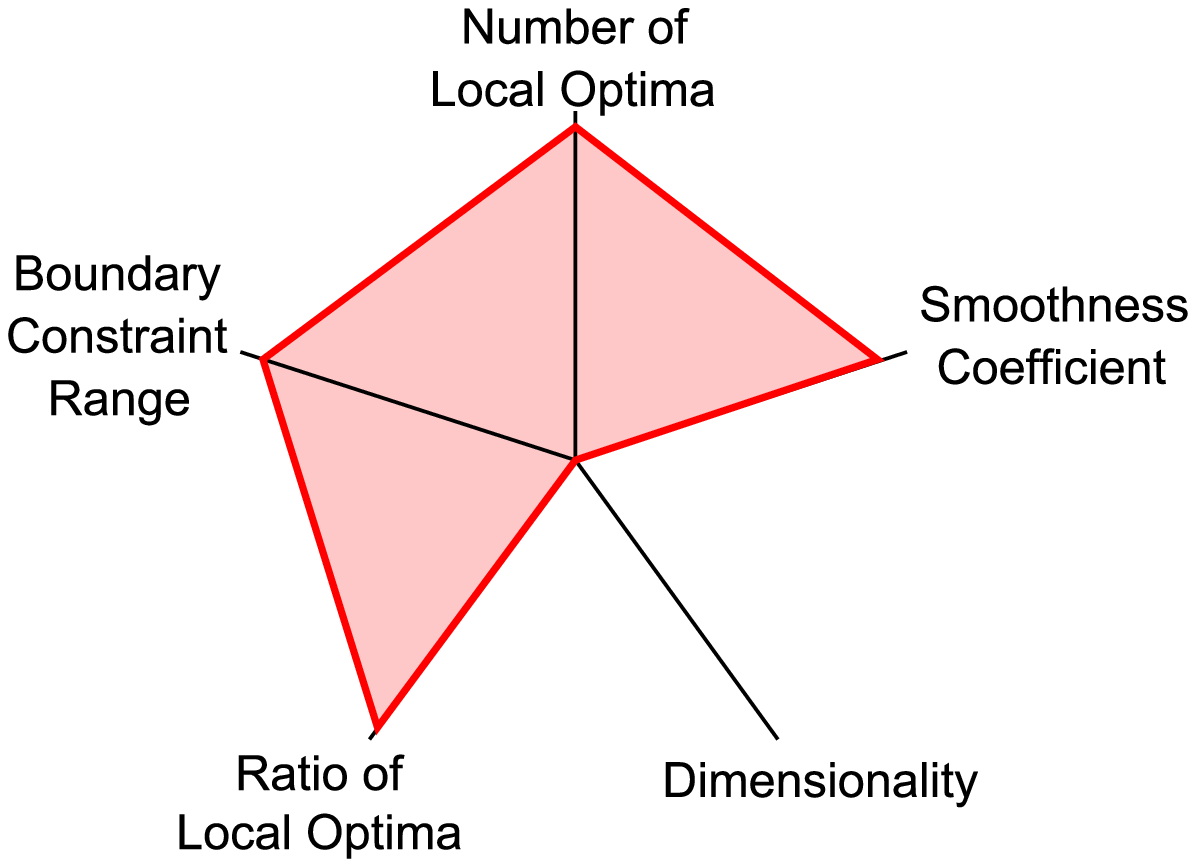}}%
			\hfil
			\subfloat[Stochastic hill climbing]{\includegraphics[height=0.2\textheight]{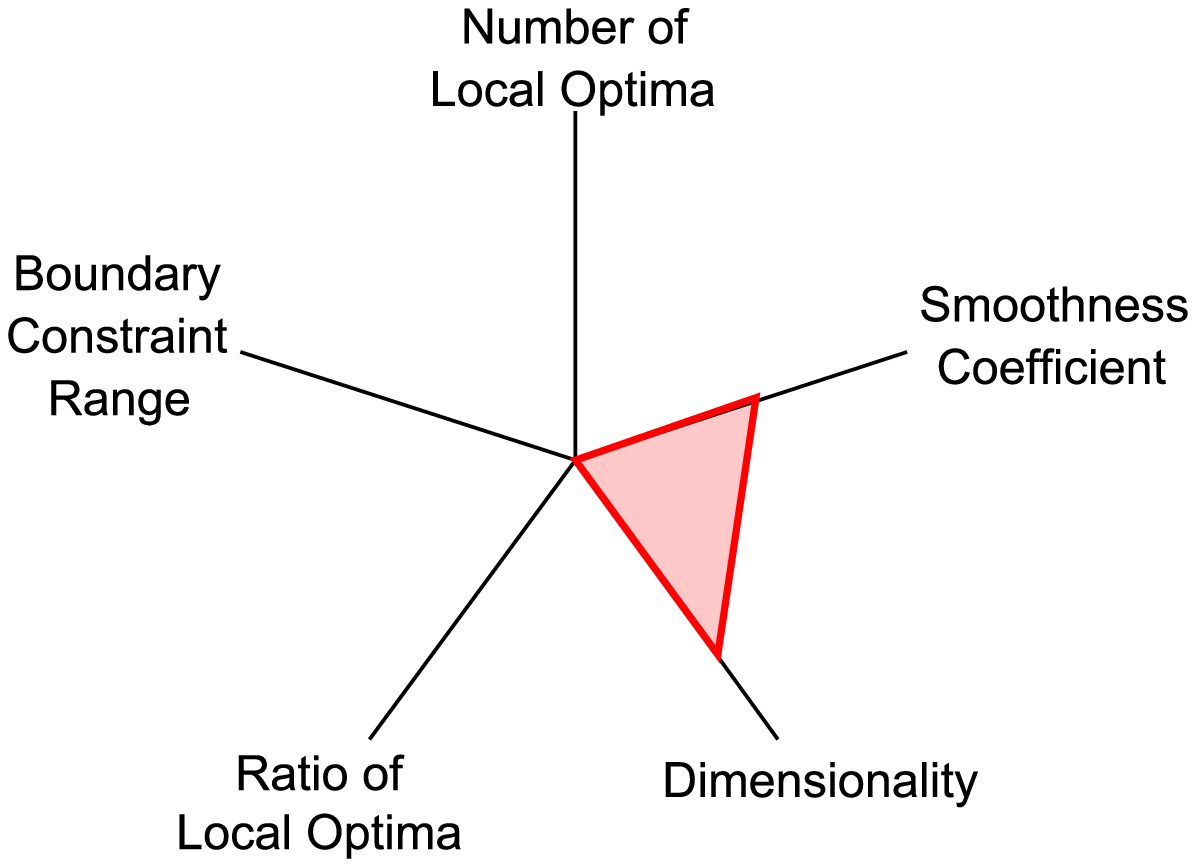}}%
			\hfil
		\caption{Radar plots depicting the standard deviation of the average error of each algorithm with respect to differing landscape characteristics.  Standard deviations are normalised on a per-axis basis.  Values close to the centre of the plot indicate a larger variance in average error, indicating these algorithms are more affected by the characteristic. In general, the more robust an algorithm, the larger the plot surface area.}
		\label{spiderplot}
	\end{figure}

Space prevents a detailed presentation of full experimental plots, but these are available from the project website{\footnote{http://www2.docm.mmu.ac.uk/STAFF/M.Amos/Project/Characterisation}}. To summarise, we plot the {\it resilience} of each algorithm to changing landscape characteristics, in the form of a radar plot in Figure \ref{spiderplot}.  To assess the resilience of an algorithm we use the standard deviation of the average error across all values of a landscape characteristic, which we normalise on a per-characteristic basis.  This ``ranking'' shows which algorithms do {\it not} show performance variability versus those which {\it are} heavily influenced by a characteristic.  BFOA shows large deviations in average error for boundary constraint range, smoothness coefficient changes and dimensionality, indicating that BFOA is an algorithm heavily dependent on the landscape of a problem - perhaps because of a heavy reliance on careful parameterisation.  SHC also shows large variance - perhaps, in large part again, to a lack of parameters and complicated local optima avoidance techniques.  GA and ES show large variation with respect to number of local optima, perhaps supporting the argument that evolutionary algorithms suffer more than most from the problem of becoming ``stuck'' in local optima.

All algorithms
produce the smallest average error when no {\bf local optima} (minima) are
present in the fitness landscape.  This is expected, as, with only one
optimum, there are no alternative solutions to which the algorithms
may converge.  We observe the greatest average error with only one
optimum from SHC, with BFOA (approx. 0.14) also showing a large
average error. There are very small average errors (almost zero) from
GA, ES, PSO, HS, RS and BA.  BFOA also produces the largest variation
in final solutions (0.32).With the introduction of only a single local optimum, performance of {\it most} algorithms degrades significantly.  ES  and GA suffer significantly, with average error increasing from approximately zero to 0.06 and 0.08 respectively, and the standard deviation of solutions increasing by around 0.15 for each algorithm. SHC also performs poorly, with a similar increase in average error.  The least affected are RS (which blindly chooses random solutions, and is therefore unaffected by local minima) and BA, which contains a global search mechanism.

For algorithms which do not directly use the gradient of the landscape, we would expect to see no change in their performance as we adjust the {\bf ratio of local optima} parameter.  We observe that RS, which selects new solutions randomly from the entire search space, offers very similar performance in terms of mean error and success rate for all ratio values.  Similarly, algorithms which perform a global search should be better at avoiding local minima even when they are attractive - and this is true for BA and HS.  PSO shows little change in success rate as the ratio becomes more attractive, owing to the fact that solutions are directed towards the best particle, and their own best solution, regardless of their individual experience with the gradation of the landscape.  Interestingly, SHC average error decreases as ratio increases - most likely due to an increased availability of `better' solutions throughout the landscape. ES demonstrates very poor, yet consistent, performance as the ratio changes.  Success rates are very low, and, interestingly, we observe a decrease in the standard deviation of solutions as the ratio increases.  This suggests that ES is perhaps more ``content" to optimise at a local minima, with the algorithm getting trapped in these more frequently as ratio increases.  This could also be true of other algorithms whose deviation decrease, such as BFOA and SHC.  GA performs in a similar manner to ES with regard to average error and diversity, although with a considerably better success rate, suggesting that this may be a general problem for algorithms which use an evolutionary approach.

At only one {\bf dimension}, fitness landscapes are trivially easy.  The performance of all algorithms reflects this, with all algorithms performing well on landscapes of a single dimension.  All algorithms show a success rate (that is, optimisation with an error of under 1.0$\times$10$^{-4}$) above 90\%.
As we increase the dimensionality to two, most algorithm performances begin to degrade.  Suffering mostly severely is RS, which is to be expected, as random search is our most basic algorithm.  Algorithms which also perform poorly at only two dimensions are ES, BA  and PSO.  It is perhaps surprising, at first, to see BA performing poorly, given that the algorithm contains a randomly sourced global search. However,  this global search is {\it effectively} RS, which performs poorly, so we can assume the global search is not covering enough of the landscape.  Coupled with the non-adaptive nature of the algorithm (meaning that solution selection around the current best area is within a relatively large range), poor algorithm performance is easily explained.  We propose that PSO and ES suffer from a similar problem, in that exploration is limited, and neither optimise their current best as accurately as their adaptive variants.  

Random search exhibits a similar, yet less extreme, reaction to changes in {\bf boundary constraints} as with the increase in dimensionality.  This is to be expected, as the limit on objective function calculations results in random search having less chance to explore the search space.  SHC also has an almost linear increase in average error, matching the linear increase in search space size, but produces consistently poor results in terms of success.  The social system algorithms (BA and PSO) both exhibit slightly unusual behaviour - as the problem space increases, their success rate also increases.  This suggests that their reliance on a parameter to search within a range is hindering the algorithms when the problem space is too small to properly explore.  HS provides the best success rate for the entire range of sizes we have selected in this problem, indicating good exploration of the search space irrespective of the range parameter.  BFOA also suffers significantly as search space size increases, again implying a heavy reliance on the parameter which controls the range of search for new solutions.  The evolutionary algorithms do not cope particularly well with the increase of problem size, with performance in terms of both average error and success rate decreasing consistently as size increases.  

The evolutionary algorithms (ES and particularly GA) perform poorly and are most affected by changing the {\bf smoothness coefficient}.  BA and PSO all also show decreasing success rate as the curves become steeper, as does BFOA which relies heavily on gradient information.
Harmony search suffers similarly to the evolutionary algorithms, and swarm algorithms, as curves become more steep.  The similarity in terms of success rate for all algorithms suggests that the availability of gradient information is something which affects all algorithms.

\section{Conclusions}
\label{conclusion}

In this paper, we have described the results of an extensive study of nature-inspired algorithms, in terms of their performance on fitness landscapes with different characteristics. We studied six nature-based methods (plus two stochastic baseline algorithms), varying a number of landscape features. The most significant characteristic appears to be the number of local minima, where a combination of global and local search appears to be beneficial. On the other hand, the ratio of local optima to the global minimum appears to have little effect on the success of the algorithms under study. As expected, dimensionality proved problematic for all algorithms, whereas landscape smoothness appeared to have little effect.

This work offers a contribution to the empirical study of nature-inspired algorithms, and we hope that it motivates future investigations. To further this work, it may be useful to examine a larger collection of nature-inspired algorithms over a greater range of values for the characteristics, in order to more fully capture a wider variety of algorithmic performance. The current work provides a firm foundation for this.  

\bibliographystyle{IEEEtran}
\bibliography{IEEEabrv,icannga_arxiv}

\end{document}